\documentclass[conference]{IEEEtran}

\usepackage{cite}
\usepackage{amsmath,amssymb,amsfonts}
\usepackage{algorithmic}
\usepackage{graphicx}
\usepackage{textcomp}
\usepackage{xcolor}
\usepackage{booktabs}
\usepackage{hyperref} 
\def\BibTeX{{\rm B\kern-.05em{\sc i\kern-.025em b}\kern-.08em
    T\kern-.1667em\lower.7ex\hbox{E}\kern-.125emX}}
\begin{document}

\title{Can Multimodal Large Language Models Understand Pathologic Movements? A Pilot Study on Seizure Semiology
}

\author{\IEEEauthorblockN{Lina Zhang}
\IEEEauthorblockA{\textit{University of California, Los Angeles} \\
linazhang@g.ucla.edu}
\and
\IEEEauthorblockN{Tonmoy Monsoor}
\IEEEauthorblockA{\textit{University of California, Los Angeles} \\
mtonmoy@g.ucla.edu}
\and
\IEEEauthorblockN{Mehmet Efe Lorasdagi}
\IEEEauthorblockA{\textit{University of California, Los Angeles} \\
lorasdagi@g.ucla.edu}
\and
\IEEEauthorblockN{Prateik Sinha}
\IEEEauthorblockA{\textit{University of California, Los Angeles} \\
prateiksinha@g.ucla.edu}
\and
\IEEEauthorblockN{Chong Han}
\IEEEauthorblockA{\textit{University of California, Los Angeles} \\
chongh3814@g.ucla.edu}
\and
\IEEEauthorblockN{Peizheng Li}
\IEEEauthorblockA{\textit{Mercedes-Benz AG} \\
peizheng.li@mercedes-benz.com}
\and
\IEEEauthorblockN{Yuan Wang}
\IEEEauthorblockA{\textit{Zhejiang University} \\
yuan2.24@intl.zju.edu.cn}
\and
\IEEEauthorblockN{Jessica Pasqua}
\IEEEauthorblockA{\textit{University of California, Los Angeles} \\
Jpasqua@mednet.ucla.edu}
\and
\IEEEauthorblockN{Colin McCrimmon}
\IEEEauthorblockA{\textit{University of California, Los Angeles} \\
CMccrimmon@mednet.ucla.edu}
\and
\IEEEauthorblockN{Rajarshi Mazumder}
\IEEEauthorblockA{\textit{University of California, Los Angeles} \\
Rmazumder@mednet.ucla.edu}
\and
\IEEEauthorblockN{Vwani Roychowdhury}
\IEEEauthorblockA{\textit{University of California, Los Angeles} \\
vwani@g.ucla.edu}
}

\maketitle

\begin{abstract}
Multimodal Large Language Models (MLLMs) have demonstrated robust capabilities in recognizing everyday human activities, yet their potential for analyzing clinically significant involuntary movements in neurological disorders remains largely unexplored. This pilot study evaluates the capability of MLLMs for automated recognition of pathological movements in seizure videos. We assessed the zero-shot performance of state-of-the-art MLLMs on 20 ILAE-defined semiological features across 90 clinical seizure recordings. MLLMs outperformed fine-tuned Convolutional Neural Network (CNN) and Vision Transformer (ViT) baseline models on 13 of 18 features without task-specific training, demonstrating particular strength in recognizing salient postural and contextual features while struggling with subtle, high-frequency movements. Feature-targeted signal enhancement (facial cropping, pose estimation, audio denoising) improved performance on 10 of 20 features. Expert evaluation showed that 94.3 percent of MLLM-generated explanations for correctly predicted cases achieved at least 60 percent faithfulness scores, aligning with epileptologist reasoning. These findings demonstrate the potential of adapting general-purpose MLLMs for specialized clinical video analysis through targeted preprocessing strategies, offering a path toward interpretable, efficient diagnostic assistance. Our code is publicly available at \url{https://github.com/LinaZhangUCLA/PathMotionMLLM}.

\end{abstract}



\section{Introduction}

Pathological movements are clinically fundamental, acting as direct diagnostic and prognostic indicators for a wide array of involuntary motor disorders \cite{martinezgarciapena2025videobased,timmermann2002mini,uchihara1994myokymia,mainka2019vocalizations}. Assessing these movements typically relies on subjective clinical observation, and manual video annotation review remains labor-intensive and time-consuming, creating bottlenecks in clinical workflows. Automated video analysis has made important strides in recent years, yet existing approaches face fundamental limitations. Most discriminative deep learning methods suffer from limited feature coverage, targeting only high-salience manifestations while neglecting subtle but diagnostically important cues. They lack interpretability, offering probability scores without explanation, a critical gap in high-stakes clinical environments \cite{amann2020xai_healthcare}. Moreover, they exhibit fragility to real-world conditions such as occlusion, lighting variation, and background noise \cite{kalitzin2012opticalflow_clonic_video}. These limitations stem from a core constraint: supervised learning requires large annotated datasets for each feature, yet expert-labeled clinical videos are scarce, and the pathological movement vocabulary is too rich to exhaustively model with fixed-label classifiers \cite{esteva2021guide_medai}.

Recent MLLMs offer a compelling alternative paradigm \cite{radford2021clip,radford2023whisper,alayrac2022flamingo,li2023blip2,liu2023llava}. Pretrained on massive web-scale corpora, MLLMs demonstrate open-vocabulary reasoning, responding to flexible natural language queries rather than predicting from fixed label sets. They generate natural language explanations that align with clinical descriptors, providing much-needed transparency. Given these capabilities, a critical question arises: Can general-purpose MLLMs, trained primarily on everyday voluntary actions, recognize the subtle, involuntary, and often ambiguous manifestations of pathological movements in real-world clinical videos?

Epileptic seizure semiology serves as an ideal representative benchmark for this pilot investigation \cite{fisher2017operational}. Seizures manifest through a diverse, temporally evolving spectrum of features, including motor, facial, autonomic, and vocal behaviors, constituting a comprehensive test for video understanding. Semiology spans from high-salience convulsions to subtle cues like oral automatisms and eye deviation. Demonstrating efficacy in seizure semiology would provide strong evidence for MLLMs' potential in broader pathological movement analysis.

This pilot study presents the first systematic evaluation of MLLMs for comprehensive seizure semiology recognition. Our contributions are threefold:

\textbf{Zero-shot MLLM Benchmarking}:
MLLMs outperformed fine-tuned CNN/ViViT baselines on 13/18 features across 20 ILAE-defined semiological features in 90 seizure videos, without task-specific training.

\textbf{Feature-Targeted Signal Enhancement}: Preprocessing strategies (facial cropping, pose overlays, audio denoising) improved performance on 10/20 features as a lightweight alternative to fine-tuning.

\textbf{Clinical Explainability Analysis}:
MLLMs generated clinically interpretable justifications with 94.3\% achieving $\geq$60\% faithfulness scores, aligning with neurologist reasoning patterns.

\section{Related Work}
\label{sec:related_work}

\subsection{Discriminative models for pathological movement understanding}

General-purpose video understanding models have made significant strides in action recognition. Spatiotemporal CNNs such as SlowFast \cite{feichtenhofer2019slowfast} employ dual-pathway architectures to capture motion at multiple temporal resolutions, while self-supervised approaches like VideoMAE \cite{tong2022videomae} learn robust representations through masked spatiotemporal prediction. VideoCLIP \cite{xu2021videoclip} further bridges vision and language by aligning video embeddings with text descriptions via contrastive learning. Despite their strong performance on everyday activities, these models are trained on general-domain action datasets (sports, daily routines) and fundamentally lack the clinical taxonomies needed to distinguish pathological movements from superficially similar voluntary behaviors.

Medical-specific vision models have emerged to address domain challenges in healthcare imaging. Models like MedViT \cite{manzari2023medvit} leverage vision transformers for tumor detection in CT scans or lesion classification in dermatology images. However, these architectures are designed for static pathology recognition and cannot capture the temporal motor dynamics that define neurological semiology. Pathological movements such as tonic-clonic seizures unfold over seconds, requiring joint modeling of spatial appearance and temporal evolution---capabilities absent in static imaging pipelines.

Seizure-specific detection systems represent the narrowest tier of existing work. Prior efforts employ specialized modules for isolated features: 3D CNNs for tonic-clonic detection \cite{boyne2025video_3dcnn_tcs}, accelerometry-based classifiers \cite{poh2012eda_accel_seizure}, and optical flow segmentation \cite{kalitzin2012opticalflow_clonic_video}. While effective for their targeted symptoms, these approaches address single features in isolation, requiring separate models for each semiological component and yielding fragmented, non-scalable solutions.

These limitations reflect three fundamental gaps in discriminative approaches for pathological movement analysis. First, existing models suffer from taxonomic misalignment: general-purpose architectures cannot natively recognize the clinically-defined vocabulary of involuntary movements without extensive task-specific retraining, while medical models remain confined to static imaging tasks. Second, the data scarcity problem remains unresolved---supervised learning demands large expert-annotated corpora that do not exist for rare neurological conditions, and current self-supervised methods have not demonstrated efficacy on pathological movement tasks. Third, all these systems lack clinical explainability: they output probability scores or binary classifications without natural language justifications, a critical barrier in high-stakes medical environments where clinicians must understand and validate automated decisions before acting on them.

\subsection{MLLMs for video understanding and medical applications}

In contrast to discriminative models, MLLMs represent a fundamentally different generative paradigm. By unifying vision encoders with large language models, MLLMs such as GPT-4V \cite{achiam2023gpt4}, Gemini \cite{geminiteam2023gemini}, and open-source alternatives like InternVL \cite{wang2025internvl3_5} and Qwen-VL \cite{bai2025qwen2_5vl} can process visual inputs with natural-language questions and generate free-form natural language descriptions. This generative capability has proven transformative for movement recognition: models can not only classify actions but also explain why a particular action was identified, describe contextual details, and respond to nuanced queries---capabilities unattainable with traditional classifiers. 

Recent MLLMs have shown substantial
promise primarily in the context of daily-life activity understanding, scene dynamics, and fine-grained motion reasoning \cite{hong2025motionbench}.  
The medical AI community has increasingly adopted MLLMs for clinical tasks, demonstrating strong performance in radiology report generation \cite{zhang2024rrg_visual_instruction}, dermatology diagnosis \cite{zhou2024skingpt4}, and medical visual question answering \cite{moor2023medflamingo}. However, these applications predominantly focus on static image analysis---interpreting chest X-rays, identifying skin lesions, or answering questions about anatomical structures. The temporal dimension remains underexplored: current medical MLLMs have not been systematically evaluated on pathological movement recognition, where clinical diagnoses depend on observing motor patterns evolving over time.

\section{Methods}
\label{sec:methods}

\begin{figure*}
    \centering
    \includegraphics[width=0.6\linewidth]{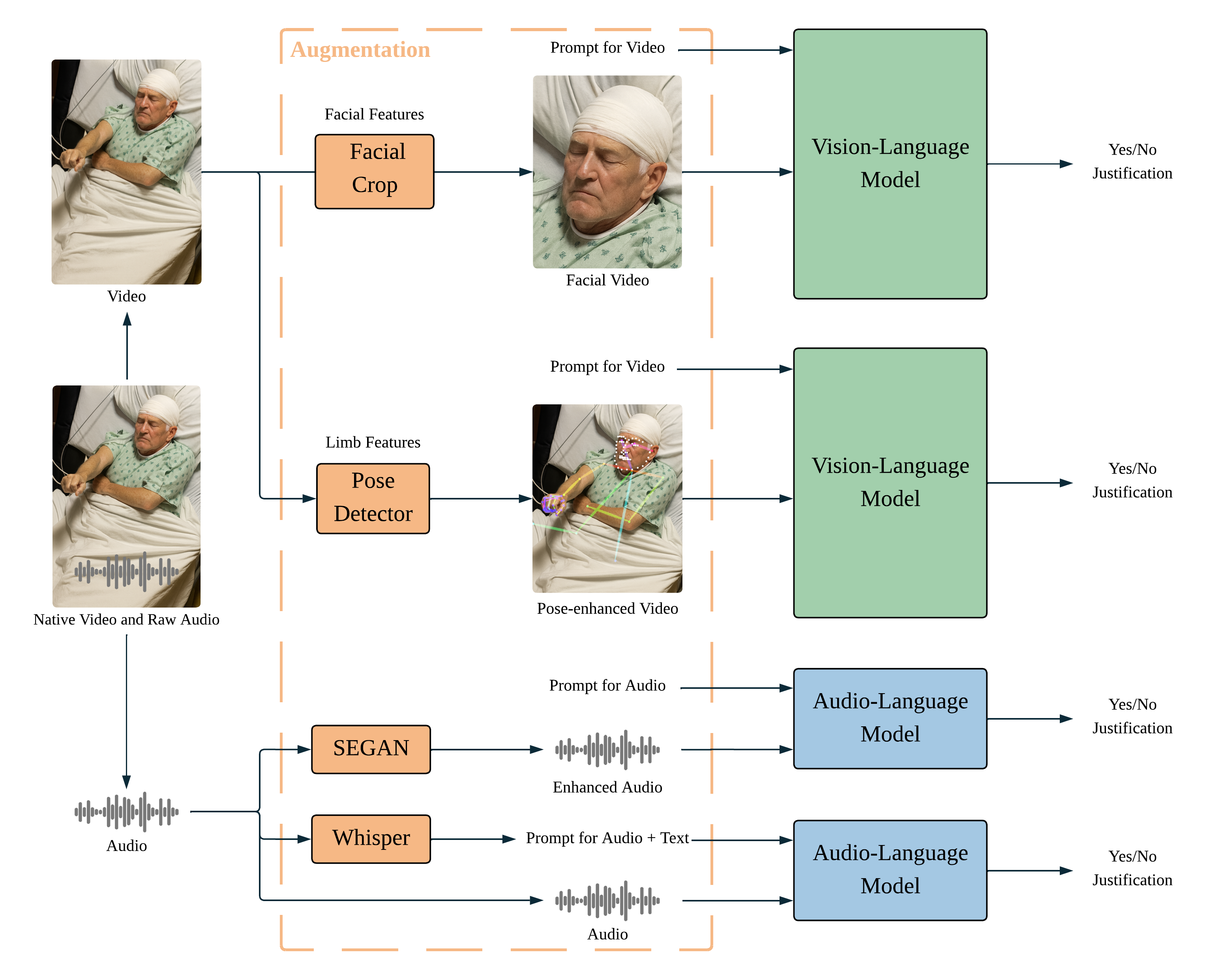}
    \caption{MLLM-based medically induced
involuntary movements recognition workflow with and without signal enhancement. (The person shown is an AI-generated virtual figure, not a real patient.)
    }
    \label{fig:workflow}
\end{figure*}

To systematically assess the potential of general-purpose MLLMs on pathological movement, we designed a comparative evaluation framework. Our methodology proceeds in two analytical stages. First, we benchmark the zero-shot performance of off-the-shelf MLLMs against task-specific, fine-tuned deep learning baselines. This comparison aims to determine whether generalist models can approximate the diagnostic accuracy of specialized supervised models without training cost. Second, we investigate if feature-targeted signal enhancement (e.g., facial cropping, pose estimation) can improve MLLMs zero-shot capabilities further (Fig.~\ref{fig:workflow}).

\begin{figure}[t]
    \centering
    \includegraphics[width=1.0\columnwidth]{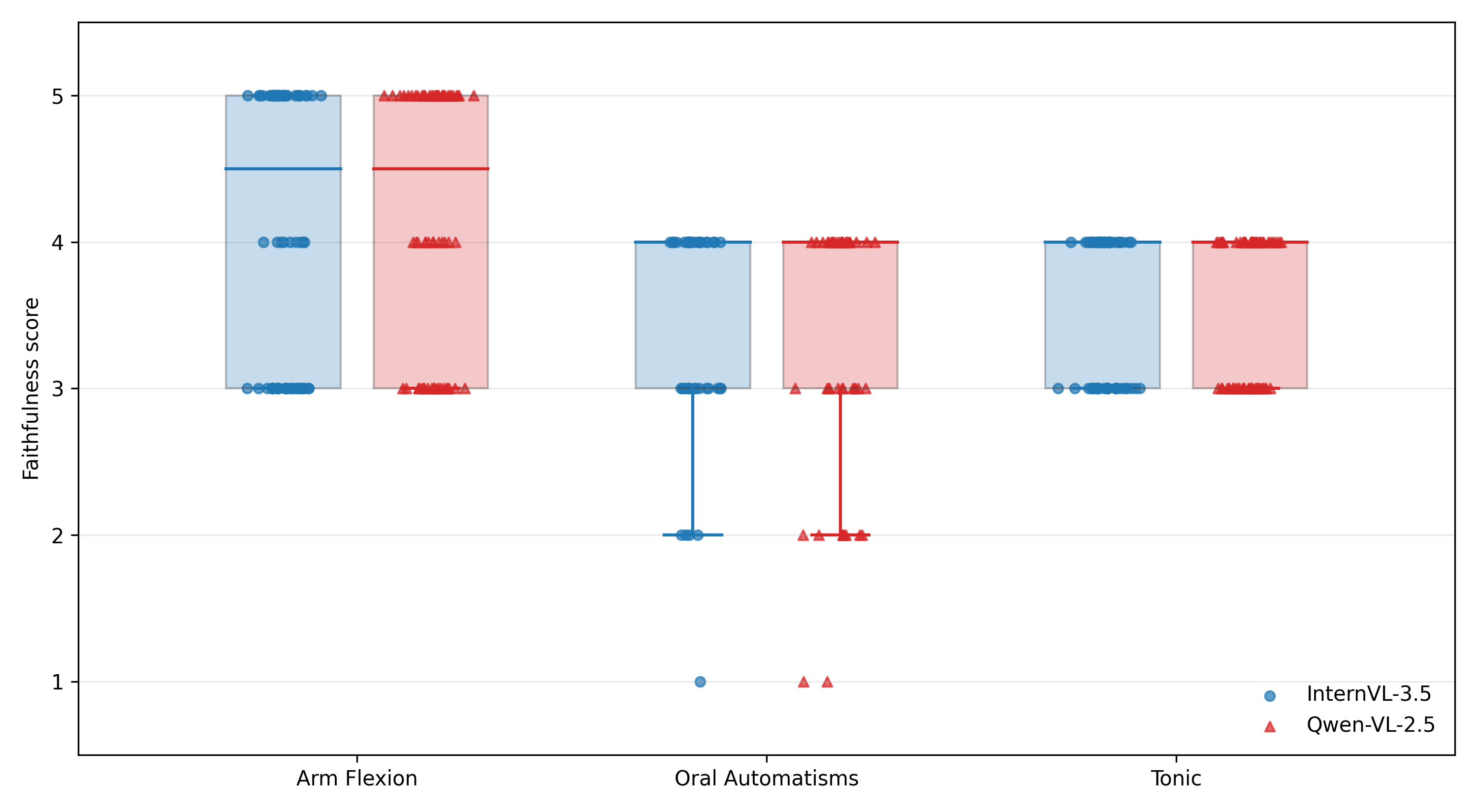}
    \caption{MLLM justification faithfulness score distribution across 3 semiological features.}
    \label{fig:score_fig}
\end{figure}

\subsection{Clinical dataset curation}

We collected 90 seizure videos from 29 consecutive adult patients undergoing video-EEG monitoring at UCLA Medical Center (2019--2023). Recordings were obtained using a fixed overhead SONY EP 580 camera with a resolution of 1920~$\times$~1080 pixels at 30 frames per second, with audio captured via unit-installed microphones at 44.1 kHz mono. To establish a robust ground truth, all videos were independently annotated by three epileptologists for the presence or absence of 20 standardized ILAE semiological motor features \cite{fisher2017operational,Fisher2017ILAEmanual}. \textit{Ethics Statement:} This study was approved by the University of California, Los Angeles Institutional Review Board under protocol IRB-23-0054.

\subsection{Direct pathological movement recognition}

Following established practices in video-based action recognition, we selected CNN \cite{tran2018closer} and Video Vision Transformer (ViViT)  \cite{arnab2021vivit} architectures as comparative baselines. Both used strong pretrained action-recognition weights: CNN from Kinetics-400 V1 and ViViT from ImageNet/Kinetics-400 \cite{carreira2017quo}. Unlike MLLMs, which we evaluate in a zero-shot manner, these supervised baselines were fine-tuned separately for each feature using clinician-provided annotations. To ensure robust evaluation and prevent information leakage, we employed a patient-stratified approach rather than a simple video-level split. Specifically, we utilized three-fold cross-validation where patients were randomly partitioned into three distinct folds. Decision thresholds were calibrated independently for each feature by maximizing the F1 score on the training folds before being applied to the held-out test fold.

We then employed state-of-the-art MLLMs to extract pathological movement features from seizure videos without any task-specific fine-tuning. 
We developed prompts collaboratively with three epileptologists, translating clinical terminology into observable behavioral descriptions. For instance, 'oral automatisms' was rephrased as 'repetitive, stereotyped mouth or tongue movements such as chewing, lip-smacking, or swallowing.' Representative prompts are shown in Table~\ref{tab:semiology}. Using these specialized prompts, we deployed InternVL-3.5 38B \cite{wang2025internvl3_5} and Qwen-VL-2.5 32B \cite{bai2025qwen2_5vl} to extract visual semiology from video segments, while Audio Flamingo 3 (AF3) \cite{goel2025audioflamingo3}  was utilized to detect auditory features from the full audio recordings.

To facilitate a direct comparison, common preprocessing and inference protocols were applied. Videos were temporally downsampled to 2 fps to balance computational efficiency with the capture of dynamic movements. During inference, each seizure recording was divided into 30-second segments, with a 5-second overlap between consecutive segments, to ensure coverage of semiological events that may span segment boundaries. Finally, segment-level detection results for both baselines and MLLMs were aggregated across the full recording using an ``any-yes'' criterion: a given semiological feature was marked as present if it was detected in at least one segment of the video.



\begin{table}[t]
\centering
\caption{Seizure semiological features and MLLM prompts samples.}
\label{tab:semiology}
\begin{tabular}{@{}p{0.32\linewidth} p{0.62\linewidth}@{}}
\toprule
\textbf{Feature} & \textbf{MLLM Prompt} \\
\midrule
Oral Automatisms & Does the patient exhibit repetitive, stereotyped mouth or tongue movements such as chewing, lip-smacking, or swallowing?   \\
Figure-4 Arms Posture & Does the patient's posture resemble a ``figure-4'' pattern, with one arm flexed and the other extended?    \\ 
Pelvic Thrusting & Does the patient display repetitive, rhythmic, anteroposterior (forward-and-backward) movements of the
hips?    \\
Ictal Vocalization & Does the patient make any groaning, moan-
ing, guttural sounds or do they utter stereotyped repetitive
phrases?   \\
\bottomrule
\end{tabular}
\end{table}

\subsection{Feature-Targeted Signal Enhancement}

Pathological movements during seizures present inherent challenges for automated detection: they often involve rapid dynamics (e.g., clonic jerks), subtle manifestations (e.g., facial twitching), and are frequently masked by clinical noise such as background conversations or medical staff interventions appearing in the camera view. These characteristics make direct feature extraction from raw videos difficult for MLLMs. Inspired by clinical practice, where neurologists instinctively focus attention on specific body regions or filter out environmental noise to better identify subtle pathological movements, we hypothesized that targeted signal enhancement could similarly guide MLLMs toward clinically salient features. 
We grouped the 20 semiological features into 3 categories (i) facial features (ii) limb features, (iii) audio features and introduced category specific  pre-processing enhancement procedures. 

\subsubsection{Facial Feature Enhancement}

For facial features, such as eye closure and facial pulling, we hypothesized that restricting the field of view to the patient's face would help the MLLM focus on clinically relevant cues. We therefore applied face detection with temporal smoothing and cropped this region before passing the frames to the MLLM.

\subsubsection{Limb Feature Enhancement}

For features, such as arm flexion, tonic, and clonic, we used a pose detector (OpenPose \cite{Cao2017OpenPose}) to identify the coordinates of the patient's key limb joints in each frame. The resulting partial-skeleton was superimposed on each frame as additional information for the MLLM.

\subsubsection{Audio Feature Enhancement}

Key auditory features, such as ictal vocalizations and verbal responsiveness, are often masked by clinical noise (e.g., alarms, conversations). To reduce interference, we integrated a SEGAN-based speech enhancement module as a front-end preprocessing step \cite{pascual2017seganspeechenhancementgenerative}. To further explore the role of contextual information, we supplemented each audio clip with its corresponding transcript extracted using OpenAI's Whisper speech recognition model (large) to convert the WAV files to text format \cite{radford2023whisper}.  The audio was treated as the primary input and the transcript was provided as secondary evidence. 

\section{Results}
\label{sec:results}

\begin{table*}[t]
\caption{Facial Features Performance: Comparison of traditional models, VLMs, and signal-enhanced VLMs.}
\label{tab:facial_features}
\centering
\renewcommand{\arraystretch}{1.2}
\setlength{\tabcolsep}{2.5pt} 
\resizebox{\textwidth}{!}{
    \begin{tabular}{l ccc ccc ccc}
    \toprule
    & \multicolumn{3}{c}{\textbf{Blank stare}} 
    & \multicolumn{3}{c}{\textbf{Closed eyes}} 
    & \multicolumn{3}{c}{\textbf{Eye blinking}} \\
    \cmidrule(lr){2-4} \cmidrule(lr){5-7} \cmidrule(lr){8-10}
    & \textit{CNN \textbar ViViT} & \textit{Qwen \textbar Intern} & \textit{Crop+Qwen \textbar Crop+Intern} 
    & \textit{CNN \textbar ViViT} & \textit{Qwen \textbar Intern} & \textit{Crop+Qwen \textbar Crop+Intern} 
    & \textit{CNN \textbar ViViT} & \textit{Qwen \textbar Intern} & \textit{Crop+Qwen \textbar Crop+Intern} \\
    \midrule
    \textbf{Accuracy}   & 0.400 \textbar 0.411 & {0.544} \textbar 0.456 & 0.533 \textbar 0.433
                        & {0.625} \textbar 0.327 & 0.535 \textbar 0.267 & 0.477 \textbar 0.267
                        & 0.747 \textbar 0.645 & 0.575 \textbar 0.770 & 0.655 \textbar 0.805 \\
    \textbf{Precision}  & 0.413 \textbar 0.421 & {0.486} \textbar 0.442 & 0.480 \textbar 0.433
                        & {0.417} \textbar 0.263 & 0.361 \textbar 0.267 & 0.317 \textbar 0.267
                        & {0.384} \textbar 0.224 & 0.074 \textbar 0.000 & 0.192 \textbar 0.000 \\
    \textbf{Recall}     & 0.922 \textbar 0.956 & 0.897 \textbar 0.974 & 0.923 \textbar {1.000}
                        & 0.514 \textbar 0.852 & 0.957 \textbar {1.000} & 0.826 \textbar {1.000}
                        & 0.444 \textbar {0.556} & 0.143 \textbar 0.000 & 0.357 \textbar 0.000 \\
    \textbf{F1 Score}   & 0.569 \textbar 0.583 & 0.631 \textbar 0.608 & \textbf{0.632} \textbar 0.605
                        & 0.410 \textbar 0.393 & \textbf{0.524} \textbar 0.422 & 0.458 \textbar 0.422
                        & \textbf{0.388} \textbar 0.314 & 0.098 \textbar 0.000 & 0.250 \textbar 0.000 \\
    \midrule
    & \multicolumn{3}{c}{\textbf{Face pulling}} 
    & \multicolumn{3}{c}{\textbf{Face twitching}} 
    & \multicolumn{3}{c}{\textbf{Oral automatisms}} \\
    \cmidrule(lr){2-4} \cmidrule(lr){5-7} \cmidrule(lr){8-10}
    & \textit{CNN \textbar ViViT} & \textit{Qwen \textbar Intern} & \textit{Crop+Qwen \textbar Crop+Intern} 
    & \textit{CNN \textbar ViViT} & \textit{Qwen \textbar Intern} & \textit{Crop+Qwen \textbar Crop+Intern} 
    & \textit{CNN \textbar ViViT} & \textit{Qwen \textbar Intern} & \textit{Crop+Qwen \textbar Crop+Intern} \\
    \midrule
    \textbf{Accuracy}   & 0.333 \textbar 0.311 & {0.611} \textbar 0.411 & 0.478 \textbar 0.489
                        & 0.433 \textbar {0.533} & 0.378 \textbar 0.378 & 0.378 \textbar 0.378
                        & 0.456 \textbar 0.444 & 0.500 \textbar {0.533} & 0.444 \textbar 0.456 \\
    \textbf{Precision}  & 0.320 \textbar 0.309 & 0.312 \textbar 0.239 & 0.295 \textbar {0.373}
                        & 0.388 \textbar {0.481} & 0.372 \textbar 0.378 & 0.378 \textbar 0.378
                        & {0.375} \textbar 0.362 & 0.354 \textbar 0.359 & 0.354 \textbar 0.333 \\
    \textbf{Recall}     & {0.926} \textbar {0.926} & 0.172 \textbar 0.379 & 0.448 \textbar 0.862
                        & 0.861 \textbar 0.699 & 0.941 \textbar {1.000} & {1.000} \textbar{1.000}
                        & {0.889} \textbar 0.833 & 0.548 \textbar 0.452 & 0.742 \textbar 0.581 \\
    \textbf{F1 Score}   & 0.463 \textbar 0.453 & 0.222 \textbar 0.293 & 0.356 \textbar \textbf{0.521}
                        & 0.531 \textbar 0.527 & 0.533 \textbar \textbf{0.548} & \textbf{0.548} \textbar \textbf{0.548}
                        & \textbf{0.524} \textbar 0.503 & 0.430 \textbar 0.400 & 0.479 \textbar 0.424 \\
    \midrule
    & \multicolumn{3}{c}{\textbf{Head turning}} 
    & \multicolumn{6}{c}{} \\ 
    \cmidrule(lr){2-4}
    & \textit{CNN \textbar ViViT} & \textit{Qwen \textbar Intern} & \textit{Crop+Qwen \textbar Crop+Intern} & \multicolumn{6}{c}{} \\
    \midrule
    \textbf{Accuracy}   & 0.667 \textbar 0.611 & {0.811} \textbar 0.800 & 0.767 \textbar 0.800 & \multicolumn{6}{c}{} \\
    \textbf{Precision}  & 0.374 \textbar 0.270 & {0.571} \textbar 0.000 & 0.364 \textbar 0.000 & \multicolumn{6}{c}{} \\
    \textbf{Recall}     & 0.417 \textbar {0.486} & 0.222 \textbar 0.000 & 0.222 \textbar 0.000 & \multicolumn{6}{c}{} \\
    \textbf{F1 Score}   & \textbf{0.325} \textbar 0.317 & 0.320 \textbar 0.000 & 0.276 \textbar 0.000 & \multicolumn{6}{c}{} \\
    \bottomrule
    \end{tabular}
}
\end{table*}

\begin{table*}[t]
\caption{Limb and Body Features Performance: Comparison of traditional models, VLMs, and signal-enhanced VLMs.}
\label{tab:limb_features}
\centering
\renewcommand{\arraystretch}{0.9}
\setlength{\tabcolsep}{2pt}
\scriptsize
\resizebox{\textwidth}{!}{%
\begin{tabular}{l ccc ccc ccc}
\toprule
& \multicolumn{3}{c}{\textbf{Occur during sleep }} & \multicolumn{3}{c}{\textbf{Arm flexion }} & \multicolumn{3}{c}{\textbf{Arms move simultaneously }} \\
\cmidrule(lr){2-4} \cmidrule(lr){5-7} \cmidrule(lr){8-10}
& \textit{CNN \textbar ViViT} & \textit{Qwen \textbar Intern} & \textit{Pose+Qwen \textbar Pose+Intern}
& \textit{CNN \textbar ViViT} & \textit{Qwen \textbar Intern} & \textit{Pose+Qwen \textbar Pose+Intern}
& \textit{CNN \textbar ViViT} & \textit{Qwen \textbar Intern} & \textit{Pose+Qwen \textbar Pose+Intern} \\
\midrule
\textbf{Accuracy}  & {0.822} \textbar 0.722 & 0.778 \textbar {0.822} & 0.778 \textbar 0.738 & 0.611 \textbar 0.611 & {0.744} \textbar 0.722 & 0.630 \textbar 0.619 & 0.518 \textbar 0.519 & {0.578} \textbar 0.278 & 0.321 \textbar 0.214 \\
\textbf{Precision} & 0.783 \textbar 0.585 & {0.933} \textbar 0.730 & 0.741 \textbar 0.600 & 0.600 \textbar 0.609 & 0.719 \textbar {0.724} & 0.597 \textbar 0.592 & 0.289 \textbar 0.254 & {0.318} \textbar 0.253 & 0.194 \textbar 0.205 \\
\textbf{Recall}    & 0.689 \textbar 0.475 & 0.424 \textbar 0.818 & 0.645 \textbar {1.000} & {0.941} \textbar 0.885 & 0.902 \textbar 0.824 & 0.881 \textbar 0.933 & 0.657 \textbar 0.567 & 0.636 \textbar {1.000} & 0.706 \textbar {1.000} \\
\textbf{F1 Score}  & 0.733 \textbar 0.510 & 0.583 \textbar \textbf{0.771} & 0.690 \textbar 0.750 & 0.731 \textbar 0.720 & \textbf{0.800} \textbar 0.771 & 0.712 \textbar 0.724 & 0.400 \textbar 0.305 & \textbf{0.424} \textbar 0.404 & 0.304 \textbar 0.340 \\
\midrule

& \multicolumn{3}{c}{\textbf{Arm straightening }} & \multicolumn{3}{c}{\textbf{Figure 4 }} & \multicolumn{3}{c}{\textbf{Tonic }} \\
\cmidrule(lr){2-4} \cmidrule(lr){5-7} \cmidrule(lr){8-10}
& \textit{CNN \textbar ViViT} & \textit{Qwen \textbar Intern} & \textit{Pose+Qwen \textbar Pose+Intern}
& \textit{CNN \textbar ViViT} & \textit{Qwen \textbar Intern} & \textit{Pose+Qwen \textbar Pose+Intern}
& \textit{CNN \textbar ViViT} & \textit{Qwen \textbar Intern} & \textit{Pose+Qwen \textbar Pose+Intern} \\
\midrule
\textbf{Accuracy}  & 0.344 \textbar 0.356 & 0.633 \textbar {0.644} & 0.580 \textbar 0.548 & 0.511 \textbar 0.789 & {0.922} \textbar 0.789 & 0.568 \textbar 0.298 & {0.444} \textbar 0.500 & 0.711 \textbar 0.711 & 0.642 \textbar 0.631 \\
\textbf{Precision} & 0.300 \textbar 0.316 & 0.442 \textbar {0.444} & 0.388 \textbar 0.353 & {0.086} \textbar 0.256 & 0.600 \textbar 0.211 & 0.135 \textbar 0.119 & 0.401 \textbar 0.367 & {0.667} \textbar 0.600 & 0.417 \textbar 0.474 \\
\textbf{Recall}    & {0.933} \textbar 0.875 & 0.852 \textbar 0.741 & 0.826 \textbar 0.783 & 0.567 \textbar 0.700 & 0.375 \textbar 0.500 & 0.625 \textbar {1.000} & 0.511 \textbar 0.847 & 0.207 \textbar 0.310 & 0.185 \textbar {0.621} \\
\textbf{F1 Score}  & 0.447 \textbar 0.442 & \textbf{0.582} \textbar 0.556 & 0.528 \textbar 0.486 & 0.126 \textbar 0.332 & \textbf{0.462} \textbar 0.296 & 0.222 \textbar 0.213 & 0.321 \textbar 0.506 & 0.316 \textbar 0.409 & \textbf{0.537} \textbar \textbf{0.537} \\
\midrule

& \multicolumn{3}{c}{\textbf{Clonic }} & \multicolumn{3}{c}{\textbf{Limb automatisms }} & \multicolumn{3}{c}{\textbf{Asynchronous movement}} \\
\cmidrule(lr){2-4} \cmidrule(lr){5-7} \cmidrule(lr){8-10}
& \textit{CNN \textbar ViViT} & \textit{Qwen \textbar Intern} & \textit{Pose+Qwen \textbar Pose+Intern}
& \textit{CNN \textbar ViViT} & \textit{Qwen \textbar Intern} & \textit{Pose+Qwen \textbar Pose+Intern}
& \textit{CNN \textbar ViViT} & \textit{Qwen \textbar Intern} & \textit{Pose+Qwen \textbar Pose+Intern} \\
\midrule
\textbf{Accuracy}  & 0.6 \textbar 0.778 & 0.667 \textbar {0.700} & 0.531 \textbar 0.548 & {0.678} \textbar 0.300 & 0.356 \textbar 0.322 & 0.395 \textbar 0.310 & 0.678 \textbar {0.700} & 0.622 \textbar 0.656 & 0.531 \textbar 0.524 \\
\textbf{Precision} & 0.293 \textbar 0.587 & 0.290 \textbar {0.333} & 0.205 \textbar 0.231 & {0.367} \textbar 0.183 & 0.224 \textbar 0.230 & 0.254 \textbar 0.256 & 0.579 \textbar {0.665} & 0.621 \textbar 0.656 & 0.433 \textbar 0.308 \\
\textbf{Recall}    & {0.808} \textbar 0.408 & 0.529 \textbar 0.588 & 0.533 \textbar 0.529 & 0.315 \textbar 0.546 & 0.714 \textbar 0.810 & 0.750 \textbar {1.000} & {0.861} \textbar 0.710 & 0.439 \textbar 0.512 & 0.382 \textbar 0.114 \\
\textbf{F1 Score}  & 0.421 \textbar 0.409 & 0.375 \textbar \textbf{0.426} & 0.296 \textbar 0.321 & 0.316 \textbar 0.274 & 0.341 \textbar 0.358 & 0.380 \textbar \textbf{0.408} & \textbf{0.690} \textbar 0.674 & 0.514 \textbar 0.575 & 0.406 \textbar 0.167 \\
\midrule

& \multicolumn{3}{c}{\textbf{Pelvic thrusting}} & \multicolumn{3}{c}{\textbf{Full body shaking }} & \multicolumn{3}{c}{} \\
\cmidrule(lr){2-4} \cmidrule(lr){5-7}
& \textit{CNN \textbar ViViT} & \textit{Qwen \textbar Intern} & \textit{Pose+Qwen \textbar Pose+Intern}
& \textit{CNN \textbar ViViT} & \textit{Qwen \textbar Intern} & \textit{Pose+Qwen \textbar Pose+Intern} & \multicolumn{3}{c}{} \\
\midrule
\textbf{Accuracy}  & 0.644 \textbar 0.589 & {0.778} \textbar 0.756 & 0.432 \textbar 0.607 & 0.598 \textbar 0.528 & {0.644} \textbar 0.556 & 0.395 \textbar 0.405 & \multicolumn{3}{c}{} \\
\textbf{Precision} & 0.312 \textbar 0.189 & 0.353 \textbar {0.370} & 0.218 \textbar 0.235 & {0.410} \textbar 0.310 & 0.318 \textbar 0.300 & 0.224 \textbar 0.242 & \multicolumn{3}{c}{} \\
\textbf{Recall}    & 0.733 \textbar 0.467 & 0.400 \textbar 0.667 & {0.800} \textbar 0.533 & 0.783 \textbar 0.733 & 0.292 \textbar 0.500 & 0.765 \textbar {0.833} & \multicolumn{3}{c}{} \\ 
\textbf{F1 Score}  & 0.423 \textbar 0.261 & 0.375 \textbar \textbf{0.476} & 0.343 \textbar 0.327 & \textbf{0.513} \textbar 0.412 & 0.304 \textbar 0.375 & 0.347 \textbar 0.375 & \multicolumn{3}{c}{} \\
\bottomrule
\end{tabular}%
}
\end{table*}

\begin{table}[t]
\caption{Audio Features Performance: Comparison of ALM (AF3) and signal enhancement with ALM.}
\label{tab:audio_features}
\centering
\renewcommand{\arraystretch}{1.15}
\setlength{\tabcolsep}{3pt} 
\resizebox{\columnwidth}{!}{%
\begin{tabular}{l ccc ccc}
\toprule
& \multicolumn{3}{c}{\textbf{Verbal responsiveness}} 
& \multicolumn{3}{c}{\textbf{Ictal vocalization}} \\
\cmidrule(lr){2-4} \cmidrule(lr){5-7}
& \textit{AF3} & \textit{Segan + AF3} & \textit{ASR + AF3} 
& \textit{AF3} & \textit{Segan + AF3} & \textit{ASR + AF3} \\
\midrule
\textbf{Accuracy}   & 0.434 & 0.321 & 0.245 & 0.765 & 0.581 & 0.744 \\
\textbf{Precision}  & 0.468 & 0.375 & 0.431 & 0.850 & 0.654 & 0.759 \\
\textbf{Recall}     & 0.361 & 0.291 & 0.327 & 0.708 & 0.500 & 0.830 \\
\textbf{F1 Score}   & \textbf{0.380} & 0.286 & 0.193 & 0.773 & 0.567 & \textbf{0.793} \\
\bottomrule
\end{tabular}
}
\end{table}

\subsection{MLLMs' Zero-Shot Ability in Detecting Seizure Semiological Features}

The best performing MLLMs outperformed the task-specific CNN and ViViT baselines (fine-tuned per feature) on 13/18 semiological features by F1 score (4/7 facial and 9/11 limb \& body). On facial semiology, MLLMs achieved clear F1 gains for \textit{closed eyes} (best baseline $F1=0.410 \rightarrow$ best MLLM $F1=0.524$), \textit{blank stare} ($0.583 \rightarrow 0.632$), and \textit{face pulling} ($0.463 \rightarrow 0.521$), and provided a smaller but consistent improvement on \textit{face twitching} ($0.531 \rightarrow 0.548$). In limb \& body semiology, MLLMs improved substantially on \textit{arm straightening} ($0.447 \rightarrow 0.582$), \textit{Figure 4} ($0.332 \rightarrow 0.462$), and \textit{tonic} events ($0.506 \rightarrow 0.537$), while also exceeding baselines on broader contextual or salient motor patterns such as occur during sleep ($0.733 \rightarrow 0.771$) and arm flexion ($0.731 \rightarrow 0.800$). Notably, the MLLM advantage was not uniform: CNN/ViViT retained higher F1 on several fine-grained or high-frequency movement features, including \textit{eye blinking} (best baseline $F1=0.388$ vs best MLLM $F1=0.250$), \textit{head turning} ($0.325$ vs $0.320$), \textit{oral automatisms} ($0.524$ vs $0.479$), \textit{asynchronous movement} ($0.690$ vs $0.575$), and \textit{full body shaking} ($0.513$ vs $0.375$).


MLLM performance patterns suggest that zero-shot generalization is strongest when the semiology is either contextual (scene-level) or visually unambiguous, and weakest when it depends on subtle, brief, or rapidly alternating motions. For example, occur during sleep reached $F1=0.771$ at $0.822$ accuracy, consistent with reliable scene understanding. Similarly, distinct motor patterns such as {arm flexion} achieved $F1=0.800$, and the best MLLM variants performed competitively on several sustained motor phenomena (e.g., {tonic} $F1=0.537$). In contrast, performance degraded for semiologies dominated by small-amplitude facial dynamics or high-temporal-frequency movements: {eye blinking} remained low even for the best MLLM configuration ($F1=0.098$), and {head turning} suffered from low or unstable precision/recall trade-offs (best MLLM $F1=0.320$ despite high accuracies up to $0.811$), indicating that these cues are often missed or confounded. Overall, these results are consistent with a zero-shot MLLM regime that is effective for coarse contextual and sustained motor signatures.

\begin{table}[t]
\centering
\caption{MLLM generated semiological feature justification samples with faithfulness score.}
\label{tab:semiology_just}
\setlength{\tabcolsep}{4pt}
\renewcommand{\arraystretch}{1.15}
\begin{tabular}{@{}p{0.24\linewidth} p{0.56\linewidth} p{0.09\linewidth}@{}}
\toprule
\textbf{Feature} & \textbf{Justification} & \textbf{Score} \\
\midrule
Oral Automatisms &
The patient is observed holding and interacting with a tablet, conversing with the nurse, and raising arms to make gestures across multiple video segments; none of these actions involve repetitive, stereotyped mouth or tongue movements indicative of oral automatisms. &
4 \\
Oral Automatisms &
The patient in the video doesn't exhibit oral automatisms. &
2 \\
Arm flexion &
The patient does not flex their arms at the elbows throughout the video segments, as they are consistently holding and interacting with a laptop. &
5 \\
\bottomrule
\end{tabular}
\end{table}

\subsection{Effect of Feature-Targeted Signal Enhancement}

Feature-targeted pre-processing proved useful, though not universally effective, in boosting MLLMs zero-shot performance. As shown in 
Tables~\ref{tab:facial_features},
\ref{tab:limb_features}, and
\ref{tab:audio_features}, enhancements improved for 10 out of the 20 semiological features.  The information added by preprocessing acts like a domain-specific attention mechanism, directing models toward clinically salient cues otherwise masked by distractors.

Facial semiological features showed most gains, and included  enhanced recognition of blank stare, blinking, face pulling, twitching, oral automatisms, and sleep-related events, though sometimes with reduced precision (e.g., blinking). Pose estimation offered a strong abstraction layer: for tonic movements, InternVL-3.5 38B's F1 rose from 0.409 to 0.537, and limb automatisms improved consistently across VLMs.

For audio, SEGAN-based denoising alone provided little benefit, likely because generative filtering altered seizure-specific sounds. By contrast, pairing transcripts with audio helped detect ictal vocalizations, slightly raising F1 (0.77 $\rightarrow$ 0.79) through improved recall. However, for verbal responsiveness, extra text reduced precision and F1, as the model sometimes misattributed background speech to the patient.

In summary, while targeted enhancements boosted recognition of several visual and auditory features, they also introduced risks, false positives, loss of context, or confusion, highlighting the need for further strengthening feature-specific preprocessing methodologies introduced in this paper.

\subsection{MLLM Explainability on Seizure Semiology}

Beyond predicting the presence or absence of semiological features, MLLMs can generate free-form natural language \emph{justifications} describing the visual and auditory cues that support their decisions. This capability is particularly relevant in epilepsy care, where clinicians routinely rely on narrative interpretations of behaviors to characterize seizure type and infer likely seizure onset networks. Table~\ref{tab:semiology_just} highlights representative explanations for multiple semiological categories, illustrating that modern MLLMs can produce descriptions that resemble clinical phrasing used in EMU reports. 

To quantitatively assess explanation quality, we conducted a structured evaluation focusing on three representative features: arm flexion, oral automatisms, and tonic movements. For each feature, expert epileptologists evaluated MLLM-generated justifications from correctly predicted samples, including all true positive cases and an equal number of randomly sampled true negative cases, using a faithfulness score ranging from 1 to 5, where each score level represents incrementally higher correctness: 1 (20\%), 2 (40\%), 3 (60\%), 4 (80\%), and 5 (100\%). Higher scores reflect justifications that are more specific, evidence-grounded, and clinically accurate in their description of observable features.

Fig.~\ref{fig:score_fig} summarizes the faithfulness score distributions. Overall, MLLMs provided reliable and clinically interpretable explanations, with $94.3\%$ justifications scoring $\geq$3 (60\% correctness or higher). However, explanation quality varied by feature type. Salient motor behaviors (arm flexion) predominantly received scores of 4-5 (median 4.5), while oral automatisms (median 3.9) and tonic (median 4) showed more scores in the 3-4 range. This disparity reflects the inherent difficulty of describing subtle facial movements compared to clear postural patterns. These findings confirm that MLLMs can provide clinically valuable explanations that support clinician-in-the-loop review, particularly for prominent motor features.

\section{Discussion and Conclusion}

This study demonstrates that general-purpose MLLMs can effectively recognize pathological movements in seizure semiology without task-specific training. Our key findings are threefold: First, zero-shot MLLMs outperformed fine-tuned CNN and ViViT baselines on 13 of 18 semiological features, achieving superior F1 scores despite requiring no domain-specific training. Second, Feature-targeted signal enhancement further improved performance on 10 of 20 features, offering a computationally efficient alternative to model fine-tuning. Techniques like pose estimation and facial cropping act as domain-specific attention mechanisms, guiding MLLMs toward clinically salient cues. Third, MLLMs provide clinically valuable explainability through natural language justifications that align with neurologist reasoning, with  $94.3\%$  explanations scoring $\geq 60\%$ faithfulness. This interpretability is critical for high-stakes clinical deployment, addressing a fundamental limitation of traditional discriminative models that only output probability scores.


We additionally performed a prompt-sensitivity analysis using Qwen2.5-VL-32B, comparing the epileptologist-designed prompts against two reduced prompt styles: simple non-expert prompts that only asked whether a visible feature occurred (e.g., ``Does this video show arm straightening?''), and concise ILAE-informed prompts that used formal semiology definitions without the additional task-specific visual descriptions used in our main prompts. For visually obvious motor features, simple non-expert prompts were competitive with the clinician-designed prompts: arm flexion achieved F1 0.824 versus the paper Qwen F1 of 0.800, and arm straightening achieved F1 0.588 versus 0.582.  Simple prompts underperformed for blank stare and closed eyes, and concise ILAE-informed definition prompts showed variable performance, including F1 0.688 for arm flexion but 0.000 for asynchronous movement, tonic, full-body shaking, and closed eyes. Some salient features are robust to prompt simplification, whereas semiology-specific terminology may be out-of-distribution for general-purpose MLLMs; expert decomposition of clinical terms into simpler observable behaviors can therefore improve performance.

The limitations of our work remain apparent. Firstly, the zero-shot accuracy for many subtle or complex features is not yet at a clinical-grade level. Secondly, our dataset, while clinically authentic, is from a single center, which may limit the generalizability of our findings. Future work should focus on two key areas: (1) Domain-specific fine-tuning of MLLMs on larger, more diverse multi-center seizure video datasets to improve their grasp of nuanced clinical cues. (2) Exploring more sophisticated methods for multimodal fusion methods, together with adaptive attention allocation mechanisms, to further enhance model accuracy and robustness.


MLLM-based systems could serve as intelligent screening assistants, automatically analyzing clinical videos and generating interpretable summaries of pathological movements. This approach significantly accelerates diagnostic workflows while maintaining transparency, establishing a practical pathway toward interpretable clinical AI.


\bibliographystyle{IEEEtran}

\bibliography{strings,refs}

\end{document}